\documentclass[acmlarge]{acmart}
\usepackage{booktabs} %

\usepackage[ruled]{algorithm2e} 

\SetAlFnt{\small}
\SetAlCapFnt{\small}
\SetAlCapNameFnt{\small}
\SetAlCapHSkip{0pt}
\IncMargin{-\parindent}
\graphicspath{ {figures/}}

\acmJournal{TKDD}
\acmVolume{0}
\acmNumber{0}
\acmArticle{0}
\acmYear{2017}
\acmMonth{0}
\acmArticleSeq{0}

\setcopyright{usgovmixed}

\acmDOI{0000001.0000001}

\received{March 2017}
\received{January 2018}
\received[accepted]{March 2018}

\begin{document}
\title{Comparison of ontology alignment systems across single matching task via the McNemar's test} 
\author{Majid Mohammadi}
\orcid{1234-5678-9012-3456}
\affiliation{%
  \institution{Delft University of Technology}
  \streetaddress{104 Jamestown Rd}
  \city{Williamsburg}
  \state{VA}
  \postcode{23185}
  \country{USA}}
\author{Amir Ahooye Atashin}
\affiliation{%
  \institution{Ferdowsi University of Mashhad}
  \department{School of Engineering}
  \city{Charlottesville}
  \state{VA}
  \postcode{22903}
  \country{USA}
}
\author{Wout Hofman}
\affiliation{%
  \institution{TNO Research Institute}
  \city{Prague}
  \country{Czech Republic}}
  
\author{Yaohua Tan}
\affiliation{%
  \institution{Delft University of Technology}
  \country{USA}}

\begin{abstract}
Ontology alignment is widely-used to find the correspondences between different ontologies in diverse fields. After discovering the alignments, several performance scores are available to evaluate them. The scores typically require the identified alignment and a reference containing the underlying actual correspondences of the given ontologies. The current trend in the alignment evaluation is to put forward a new score (e.g., precision, weighted precision, semantic precision, etc.) and to compare various alignments by juxtaposing the obtained scores. However, it is substantially provocative to select one measure among others for comparison. On top of that, claiming if one system has a better performance than one another cannot be substantiated solely by comparing two scalars. 
In this paper, we propose the statistical procedures which enable us to theoretically favor one system over one another. The McNemar's test is the statistical means by which the comparison of two ontology alignment systems over one matching task is drawn. The test applies to a $2 \times 2$ contingency table which can be constructed in two different ways based on the alignments, each of which has their own merits/pitfalls. The ways of the contingency table construction and various apposite statistics from the McNemar's test are elaborated in minute detail. In the case of having more than two alignment systems for comparison, the family-wise error rate is expected to happen. Thus, the ways of preventing such an error are also discussed. A directed graph visualizes the outcome of the McNemar's test in the presence of multiple alignment systems. From this graph, it is readily understood if one system is better than one another or if their differences are imperceptible. The proposed statistical methodologies are applied to the systems participated in the OAEI 2016 anatomy track, and also compares several well-known similarity metrics for the same matching problem. 

\end{abstract}

%
%


%
%



%

\maketitle

\renewcommand{\shortauthors}{M. Mohammadi et al.}

\section{Introduction}
With the advancement in information technology, data these days come from various sources. Such data have multiple salient but unwelcome features: they are big, dynamic and heterogeneous. There are solutions to cope with any of these features, and ontology alignment (or mapping/matching) is a remedy to data heterogeneity \cite{om_book}. 

Given the source and target ontologies for alignment, a correspondence is defined as the mapping of one concept in the source to one concept in the target ontology. For discovering correspondences, it is typical to utilize one or more similarity measures. There are three different categories for the similarity calculation \cite{om_book}. The first category is the string-based measures which only considers the \textit{text} of concepts to compute their similarities \cite{smoa,levenstein,st_3}. Another group is the linguistic-based similarity measures which consider the linguistic relations, e.g. synonym, antonym, hypernym, etc., between the strings of two concepts. The linguistic-based similarity measures usually take advantages of WordNet \cite{wordnet} to discover the similarity. The third class is the structural-based measures which take into account the position of concepts in their ontologies.

Yet another approach is to match the entities of two given ontologies based on their instances \cite{instance_matching}. The underlying assumption behind this type of alignment is that two entities are similar provided that they share, more or less, analogous instances. 

Traditionally, the challenge of ontology alignment was to come up with a new similarity measure and then to find the interrelation between the ontologies \cite{smoa}. However, this focus has moved to take advantages of various similarity measures and try to reason correspondences based on the outcomes of various metrics \cite{rough,dssim}.

An alignment, which is the result of any standard ontology matching system, comprises a set of correspondences, mapping various concepts of one ontology to those of the other. It is the common practice to find the goodness of an alignment system by comparing its output with the actual reference alignment which is in hand. The typical performance scores are the precision and recall along with their variation such as relaxed precision and recall \cite{relaxed_pr}, semantic precision and recall \cite{semantic_pr}, and so on. However, it is controversial to select the appropriate performance score in different cases. For instance, the comparison based on precision and recall would lead to totally different results. A system can be quite precise and discover as few false correspondences as possible, e.g., high precision, but could be conservative and not be able to detect an acceptable portion of correspondences, e.g., low recall. In addition to the selection of a performance metric, claiming the superiority of a system against one another cannot be substantiated merely by comparing the acquired scores. The difference between the performance measures of two systems could be small and imperceptible, thereby asserting the superiority of one system might not be correct. One approach to support such allegations and verify if the difference between two systems is substantial would be the statistical analysis. In this article, the appropriate procedures are put forward to statistically opt for one system if it has an enhanced performance than the other. 

A note of caution is in order at this point, however. According to the no free lunch theorem \cite{nfl,nfl2}, there is no context-independent reason to favor one strategy (or optimization method) over one another, and the average performances of all strategies over all possible problems are the same. It is drawn, as a result, that the superior performance of one method over one another is due to its better fitness to the nature of the problem, not because of its inherent features. Any claim of performing \textit{the best} in a general sense must be questioned and faced with doubts.

The no free lunch theorem is firstly introduced in the supervised machine learning realm \cite{nfl3}, but it is generalized to any optimization problem afterward \cite{nfl}. Therefore, the results of the no free lunch theorem are also correct for the ontology matching problem, and the preferred alignment can be only recognized in one particular context.

To date, the attempt of claiming if one alignment system is better than one another has been solely concentrated on employing a new performance score, e.g. semantic precision, relaxed precision, etc. \cite{relaxed_pr,semantic_pr,precisionN}. If there are multiple pairs of ontologies for comparison, the superiority of a system is dedicated only if its average performance across multiple pairs of ontologies is higher than the rest. Statistically speaking, the average performance is unsafe and inappropriate: it is highly sensitive to outliers and having higher average performance does not necessarily indicate the superiority since the difference might be imperceptible and insignificant \cite{demsar}. In the case of existing only one pair of ontologies, on the other hand, the comparison is merely performed by the juxtaposition of the performance metric of various systems.

As a complement to the no free lunch theorem, this article aims to consider the statistical hypothesis testing to find the best ontology alignment on a particular task. Employing the appropriate statistical test, one can determine if one alignment system outperforms one another with  substantial statistical evidence. Instead of comparing one alignment with the reference one, the recommended methodology here takes the reference along with two alignments under comparison as the inputs and states if one of them statistically outperforms the other. Thus, the expected outcome is not a score but the statement of superiority of an alignment in comparison with one another.

In the case that there are multiple tasks, various statistics such as Wilcoxon signed-rank and Friedman tests can be applied to a particular performance score obtained for each matching task \cite{omstatistical}. In order words, the performance scores obtained from each task become the samples, hence the difference between systems can be gauged by conducting statistical tests over the samples. However, it is not the case for comparison over one matching task since there is no such samples.

The McNemar's test is the statistical means by which the various matching systems can be compared over one matching task. This test can be applied to the paired nominal data summarized in a contingency table with a dichotomous trait. Interestingly, the outcome of two alignment systems can be viewed as dichotomous (i.e., correct and incorrect correspondences) of two experiments (i.e., two alignment systems). Therefore, the McNemar's test suits for comparison of alignments. However, Summing up the results of alignments in a contingency table would be challenging and might erupt discussions. We present two ways to build such a contingency table whose applicabilities is conceptually similar to those of recall and F-measure. Further, four statistics from the McNemar's tests are considered, and their advantages and pitfalls are discussed. In the case of having two systems for comparison, the McNemar's test can be simply applied. If more than two alignments are available, all pairwise comparisons must be performed. In this case, the family-wise error rate (FWER) is likely to happen and must be controlled \cite{fwer}. The appropriate procedures for the FWER prevention are elaborated as well. 

We leverage the proposed methodology across the systems participated in the OAEI 2016 \textit{anatomy} track, and the corresponding results are visualized by a directed graph. This graph indicates if the difference between each pair of systems are significant or not. Our investigation shows that AML and CroMatcher are the top two systems while DKP-AOM and Alin are the ones with reduced accomplishment. We further compare the string-based similarity measures over this track because many correspondences can be easily discovered by comparing the strings. The N-gram and Levenstein distances are the ones with the maximum discovery with respect to others.

The contribution of this paper can be summarized as
\begin{itemize}
\item The utilization of the McNemar's test to conduct the comparison of alignment systems.
\item Two ways of using the McNemar's test is proposed which are conceptually identical to those of recall and F-measure. 
\item The technique for the family-wise error rate is thoroughly discussed.
\item The outcome of the statistical procedure for comparison of multiple systems is visualized by a directed graph.
\item The systems participated in the OAEI anatomy track are compared and the corresponding results are reported.
\end{itemize}

This article is structured as follows. The ways of the contingency table construction are expounded in Section II, and the appropriate statistics from the McNemar's test are discussed in Section III. The family-wise error rate and the ways of adjusting the p-values are studied in Section IV. Section V dedicates to the experiments of the statistical procedures over the anatomy track, and the paper is concluded in Section VI.

\section{Contingency table construction}
The McNemar's test is applicable when there are two experiments over N samples \cite{mcnemar}. Let the outcome of each test be either positive or negative; then, a simple contingency table would be as Table \ref{tab:contingency}.

\begin{table}[h!]
\caption{A simple contingency table}
\label{tab:contingency}
\centering
\small
\begin{tabular}{c c c c c }
\hline
	   &      & \multicolumn{2}{c}{Exp. 2} & \\ 
	   &      &    -     &   +      & sum  \\ \hline
Exp. 1 &  -   & $n_{00}$ & $n_{01}$ & $n_{0.}$ \\
       &  +   & $n_{10}$ & $n_{11}$ & $n_{1.}$ \\
       & sum  & $n_{.0}$ & $n_{.1}$ & N \\ \hline
\end{tabular}
\end{table}
In this table, $n_{00}$ and $n_{11}$ are called the accordant pair and are respectively the number of times both experiments produce positive and negative outcomes. The discordant pair, i.e. $n_{01}$ and $n_{10}$, are the number of times the results of experiments are in contradiction; $n_{01}$ is the number of experiments which the first outcome is negative while the second one is positive and $n_{10}$ is the other way around.

In the ontology matching case, the positive or negative outcome can be defined in two ways, each of which has its own merits and is suitable for particular situations.

For two given ontologies, let $R$ be the reference alignment containing a set of correct correspondences and $A_1$ and $A_2$ be two alignments retrieved by two different systems. In the first approach of the contingency table construction, the focus is solely on the truly discovered alignments, thereby ignoring the concepts which have not correctly mapped. Hence, $n_{00}$ and $n_{11}$ are respectively the number of false correspondences and the number of correct correspondences jointly identified by both systems. $n_{01}$ (and similarly $n_{10}$) is the number of correspondences correctly discovered by $A_2$, but not by $A_1$. These elements can be written as
\begin{align}\label{ct_nfp}
\centering
\begin{cases}
&n_{00} = \vert R - (A_1 \cup A_2) \vert\\
&n_{01} = \vert (A_2 \cap R) - A_1\vert \\ 
&n_{10} = \vert (A_1 \cap R) - A_2 \vert \\ 
&n_{11} = \vert A_1 \cap A_2 \cap R\vert \\
\end{cases}
\end{align}
where $|.|$ indicates the cardinality operator. This approach is conceptually similar to $recall$ as it does not consider the wrong correspondences in the alignments. We again accent that the approach of this article is distinct from the performance measures, including recall, as we compare two alignments and do not produce any score indicating the fineness of a system.

An example elaborates the issue of this approach. Assume that two systems could discover the complete reference alignment, i.e. $A_1 = A_2 = R$. In this case, $n_{01} = n_{10} = 0$ which means that they are equally well (it is discussed in further sections that $n_{01}$ and $n_{10}$ are the only important pair for the McNemar's test). Now, suppose that $A_1 = R$ and $A_2 = R+B$, where $B$ is a set of correspondences which are not in R (falsely discovered by $A_2$). In this case, $n_{01}$ is the same as $n_{10}$ which again indicates that their performances are indiscernible. However, it is plain to grasp that $A_1$ is more reliable as it does not mistakenly discover any correspondences. Statistically speaking, this approach does not take into account the false positive and only considers the true positive. Nonetheless, such an approach is suitable for occasions where the goal is to have as many correspondences as possible so that the false discovery does not have a profound impact.

The second approach of building the contingency table avoids the foregoing pitfall and consider the false discovery as well. Since it considers the truly unmapped pairs of concepts, obtaining the elements of the contingency table is of higher complexity in comparison with the previous approach. Therefore, it is necessary to explain how to obtain each element of the table individually.

$n_{00}$ is the number of correspondences which are wrongly discovered by both alignments. Hence it includes the correspondences which are in R but not in $A_1$ or $A_2$ plus the correspondences which are in both $A_1$ and $A_2$ but not in R, i.e. $n_{00} = \vert R - (A_1 \cup A_2)\vert + \vert (A_1 \cap A_2) - R \vert$. $n_{10}$ is the number of truly discovered correspondences by $A_1$ which are not in $A_2$ plus the correspondences which are falsely identified only by $A_2$ and not by $A_1$, i.e. $n_{10} = \vert(A_1 \cap R) - A_2\vert + \vert A_2 - A_1 - R\vert$. By the same token, $n_{01}$ can also be obtained. $n_{11}$ is a bit more challenging as the total number of possible correspondences between two ontologies is required. Let this number be $T$, one possibility for $T$ is to multiply the number of concepts of two ontologies, i.e. $T = n \times m$ where $n$ and $m$ are the numbers of candidate concepts for matching in two ontologies. Thus, $n_{11} = \vert A_1 \cap A_2 \cap R\vert + \vert (T - R) - (A_1 \cup A_2)\vert$. The statistics considered in this paper only need the discordant pair; therefore the value of $n_{11}$ and subsequently $T$ is not taken into account. The elements as mentioned earlier of the contingency table from the second approach can be summarized as:
\begin{align}\label{ct_wfp}
\begin{cases}
&n_{00} = \vert R - (A_1 \cup A_2)\vert + \vert (A_1 \cap A_2) - R \vert \\
&n_{01} = \vert(A_2 \cap R) - A_1\vert + \vert A_1 - A_2 - R \vert \\ 
&n_{10} = \vert(A_1 \cap R) - A_2\vert + \vert A_2 - A_1 - R \vert \\ 
&n_{11} = \vert A_1 \cap A_2 \cap R\vert + \vert (T - R) - (A_1 \cup A_2)\vert \\
\end{cases}
\end{align}

This way of the contingency table construction considers the false correspondences as well. The foregoing example illustrates the advantages of these formulas. As $A_1 = R$ and $A_2 = R+B$, $n_{01} = 0$ and $n_{10} = |B|$. The null hypothesis is thus rejected for large enough of B, and A is claimed to be superior. Therefore, the false positive of B resulted in declaring A to be the better system. Note that this calculation is relative to the other system. In other words, it does not consider all the incorrectly identified correspondences, but the false correspondences are computed as the ones which are not in the rival system. As the goal is to compare two alignments together, it is entirely logical to find the \textit{relative false positive}. This approach can be figuratively viewed as similar to F-measure due to its consideration of both true and false discoveries. 

\section{McNemar's test}

The McNemar's test is applied to the contingency table constructed in the previous section. But before looking into the test, we digress briefly to explain the null hypothesis testing.

To leverage any statistical test, the null and alternative hypotheses are required. The null hypothesis $H_0$ states that the difference between two populations is insignificant, and the existing discrepancy is due to the sampling or experimental errors \cite{handbook}. The alternative hypothesis, on the other hand, states the contrary: the difference between two populations is significant and not random.

To reject or retain $H_0$, we need to compute the p-value and compare it with significant level $\alpha$ which must be determined before running the test. The p-value is the probability of obtaining a result equal to, or even more extreme than, the observations given the null hypothesis is true \cite{handbook}. If the p-value is less than the nominal significant level $\alpha$, then the null hypothesis is rejected, and it is drawn that the  disparity between populations is significant.

In comparison of ontology alignment systems, the populations mentioned above are the outcomes of two systems. Therefore, the null hypothesis is that the difference between the outcomes of alignments is random and insignificant. The null hypothesis in the McNemar's test states that the two marginal probabilities of the contingency table are the same, i.e.
\begin{align}\label{mcnemar_homogeneity}
p(n_{00}) + p(n_{01}) = p(n_{00}) + p(n_{10}) \cr
p(n_{10}) + p(n_{11}) = p(n_{01}) + p(n_{11})
\end{align}
where $p(a)$ indicates the probability of occurring the cell of Table \ref{tab:contingency} with the label \textit{a}. After canceling out the $p(n_{00})$ and $p(n_{11})$ from the foregoing equations, the null and alternative hypotheses become
\begin{align}\label{mcnemar_hypo}
&H_0: \quad p(n_{01}) = p(n_{10}) \cr
&H_a: \quad p(n_{01}) \neq p(n_{10}).
\end{align}
To compute the p-value of the null hypothesis \eqref{mcnemar_hypo}, we consider four statistics from the McNemar's test and discuss their advantages and pitfalls in the hypothesis testing. The statistics studied here only work with the accordant pair of the contingency table. However, there is also an exact unconditional McNemar's test which takes into account the discordant pair of the contingency table \cite{mcnemar_uncon}. The exact unconditional test is way more intricate than the McNemar's tests put forward here, but its power is  approximately the same as other tests \cite{mcnemar_all}. Therefore, this test is ignored in this paper.

\subsection{The McNemar's asymptotic test}
The McNemar's asymptotic test assumes that $n_{01}$ is binomially distributed with $p=0.5$ and parameters $n = n_{01}+n_{10}$ under the null hypothesis \cite{mcnemar}. The McNemar's asymptotic statistic 
\begin{align*}
\chi^2 = \frac{(n_{01} - n_{10})^2}{n_{01} + n_{10}}
\end{align*}
is distributed according to $\chi^2$ with one degree of freedom. This test is undefined for $n_{01} = n_{10} = 0$.

To reject the null hypothesis, this test requires a sufficient number of data ($n_{01} + n_{10} \geq 25$) since it might violate the nominal significant level $\alpha$ for the small sample size.

\subsection{The McNemar's exact test}
It is traditionally advised to use the McNemar's exact test when a small sample size is available in order not to exceed the nominal significant level. In this test, $n_{01}$ is compared to a binomial distribution with parameter $n = n_{01} + n_{10}$ and $p = 0.5$. Thus, the p-value for this test is obtained as
\begin{align*}
\text{exact-p-value} = \sum_{x=n_{01}}^n \begin{pmatrix}n\\ x\\\end{pmatrix} \left(\frac{1}{2}\right)^2
\end{align*}
The two-sided p-value is calculated by multiplication of the one-sided p-value by two. This test guarantees to have type I error rate below the nominal significant level $\alpha$. 

\subsection{The McNemar's asymptotic test with continuity correction}
The main drawback of the McNemar's exact test, though preserving the nominal significant level, is conservatism: it unnecessarily generates large p-values so that the null hypothesis cannot be rejected. As a remedy to conservatism, Edwards \cite{edwards} approximated the exact p-value by the following continuity corrected statistic
\begin{align*}
\chi^2 = \frac{(|n_{01}-n_{10}|-1)^2}{n_{01} + n_{10}}
\end{align*}
which is $\chi^2$-distributed with one degree of freedom. This test is also undefined for $n_{01} = n_{10} = 0$.

\subsection{The McNemar's mid-p test}
The continuity corrected method is not as conservative as the exact test, but it does not guarantee to preserve the nominal significant level. The mid-p approach propounds a way to trade off between the conservatism of the exact tests and the significant level transgression of the continuity correction approach \cite{midp}. To obtain the mid-p-value, a simple modification is required: the mid-p-value equals the exact p-value minus half the point probability of the observed test statistic \cite{mcnemar_all}. Hence, the p-value could be computed as
\begin{align*}
\text{mid-p-value} = \text{2-sided exact p-value} - \begin{pmatrix}n\\ n_{01}\\\end{pmatrix}0.5^n.
\end{align*}
The McNemar's mid-p test resolves the conservatism of the exact test, but it does not guarantee theoretically to preserve the nominal significant level. In a recent study, however, it is investigated that the mid-p test has low type I error and does not violate the significant level. The continuity-corrected test, in contrast, indicated a high type I error, coming from the nature of asymptotic tests, as well as high type II error, inherited from the exact test. Thus, it is rational not to use the continuity-corrected test for the alignment comparison.

\section{Family-wise error rate and p-value adjustment}
When there are two systems for comparison, the null hypothesis will be rejected if the obtained p-value is below the nominal significant level $\alpha$. If more than two alignments are available for comparison, the well-known family-wise error rate (FWER) might occur. FWER refers to the increase in the probability of type I error which is likely to violate the nominal significant level $\alpha$ when multiple populations are to be compared. 
To explain what FWER is, assume that there are $5$ systems for comparison and the significant level is $\alpha = 0.05$. If it is desired to do all the pairwise comparisons, then there are $k = 5\times 4 / 2= 10$ hypotheses overall. For each of null hypotheses, the probability of rejection without occurring the type I error is $1- \alpha = 0.95$. For all comparisons, on the other hand, the probability of not having any type I error in all the hypotheses is $(0.95)^{10} = 0.6$. As a result, the probability of occurring at least one type I error increases to $1 -  0.6 = 0.4$, which is way higher than the nominal $\alpha = 0.05$. This phenomenon is the so-called family-wise error rate.

To prevent this error, there are two primary approaches. Akin to the preceding example, the first approach is applicable when all the pairwise comparisons are desired. Conducting all pairwise comparisons are suitable when a comparison study of the existing systems in the literature or their competition in a competition like OAEI is desired. Another approach to control FWER is convenient when a new alignment system is proposed and it is to be compared with other existing ones. In the interest of simplicity, the former approach is called $N\times N$ comparisons and the latter is called $N\times 1$ comparisons.

\subsection{Controlling FWER in $N\times 1$ comparison}
When a new alignment system is proposed, it is usually compared with other existing ontology matchers. For comparing $n$ systems (including the proposed one) in this case, $k=n-1$ comparisons must be performed. There are four methods which can control the family-wise error rate in this case. These methods can be viewed as the p-value adjustment procedures which modify the p-values in a way that the adjusted p-values (APV) can be directly compared with the significance level while the nominal significant level is also preserved. Thus, a null hypothesis is rejected if its corresponding adjusted p-value is below the nominal $\alpha$.

Let $H_i, i=1,...,k$ be all hypotheses for $n$ systems and $p_i, i=1,...,k$ be their corresponding p-values. The Bonferroni's method \cite{bonferroni} is the most straightforward way to prevent FWER. In this procedure, all the p-values are compared with the nominal significant level $\alpha$ divided by the total number of comparisons. In other words, the hypothesis $H_i$ is rejected if $p_i < \alpha / k$. Based on this equation, the adjusted p-value for the hypothesis $H_i$ is obtained by multiplying both sides of above inequality by $k$, i.e. $APV_i = \min\{k\times p_i,1\}$. Thus, $H_i$ is rejected if $APV_i < \alpha$. This procedure, though simple, is too conservative: it retains the hypotheses which must be rejected by generating high APV.

In contrary to the single step Bonferroni's correction, there are step-up and step-down procedures which sequentially reject the null hypothesis. It is necessary to order p-values for sequential rejective procedures and we denote the ordered p-values as $p_1 \leq p_2 \leq ...\leq p_k$ and their corresponding hypotheses as $H_1,H_2,...,H_k$.

The Holm's procedure \cite{holm} is a step-down method which starts with the most significant (or the smallest) p-value $p_1$. If $p_1 \leq \frac{\alpha}{k}$, then $H_1$ is rejected, and $p_2$ is compared with $\frac{\alpha}{k-1}$. If $p_2 \leq \frac{\alpha}{k-1}$, then $H_2$ is rejected, and $p_3$ is compared with $\frac{\alpha}{k-2}$. This procedure continues until a hypothesis is retained. In other words, each $p_i$ in the Holm's method is compared with $\frac{\alpha}{k+1-i}$ and it is rejected if it is below this value; otherwise, it is not rejected and the rest hypotheses are retained as well. The Holm's adjusted p-value is $APV_i = \min\{v_i,1\}$ where $v_i = max\{(k-j)p_j : 1 \leq j \leq i \}$. 

Similar to the Holm's procedure, the Holland's correction \cite{holland} is also a step-down method. Instead of comparing the p-values with $\frac{\alpha}{k+1-i}$, it compares each $p_i$ with $1-(1-\alpha)^{k-i}$. Thus, the adjusted p-value is $APV_i = \min\{v_i,1\}$ where $v_i = max\{1 - (1-p_j)^{k+1-j}: 1 \leq j \leq i \}$. The Finner's procedure \cite{finner} is almost the same as the Holland's technique and compares each $p_i$ with  $1-(1-\alpha)^{\frac{k}{i}}$. The Finner's adjusted p-value is $APV_i = \min\{v_i,1\}$ where $v_i = \max\{ 1 - (1-p_j)^{\frac{k}{j}}: 1 \leq j \leq i \}$.

The Hochberg's method \cite{hochberg} works in the opposite direction and starts with the largest p-value. It compares the largest p-value with $\alpha$, the next largest with $\alpha / 2$ and it is terminated until a hypothesis is rejected. All the hypotheses with the smaller p-values are then rejected as well. The Hochberg's adjusted p-value is $APV_i = \max\{(k-j)p_j: (k-1)\geq j \geq i \}$.  \\
 
\subsection{Controlling FWER in $N\times N$ comparison}
For performing all the pairwise comparisons when $n$ systems are available, there are $k = n(n-1)/2$ hypotheses overall. The Nemenyi's method \cite{nemenyi} is exactly the Bonferroni's correction with $k$ is set to the $N\times N$ comparison, i.e. $k = n(n-1)/2$. Thus, it has high type II error which results in not detecting the difference among the population when there is a de facto difference.
The same modification of $k$ must be applied to other methods so that they are suitable for $N \times N$ comparison case.

There is also another sequential-rejective null hypothesis approach which is suitable for $N\times N$ comparison. This approach takes into account the logical relations between hypotheses. Shaffer \cite{shaffer} discovered that the Holm's procedure could be improved when hypotheses are logically interrelated. In many scenarios, it is not feasible to get any combination of true and false hypotheses. In the pairwise comparison, for instance, it is not possible to have $\mu_1 = \mu_2$ and $\mu_2 = \mu_3$ but $\mu_1 \neq \mu_3$. Thus, this case need not be protected against FWER.

Correction procedures which take into account the logical relations are similar to the Holm's correction: they start with the most significant (or the smallest) p-value but compare it with $\alpha / t_1$, where $t_1$ is the maximum number of hypotheses which can be retained at the first step. If $p_1 < \alpha / t_1$, then the corresponding hypothesis $H_1$ is rejected, and $p_2$ is compared with $\alpha / t_2$. If $H_2$ is rejected, then $p_3$ is compared with $\alpha / t_3$ and so on. The procedure terminates at the stage $j$ if $H_j$ cannot be rejected. The remaining hypotheses with bigger p-values than $p_j$ are also retained. The adjusted p-value for the sequential corrective methods is $APV_i = \min\{v_i,1\}$ where $v_i = \min\{t_i\times p_i,1\}$.

There are two well-known techniques which consider the logical relations of hypotheses: Shaffer's and Bergmann's. These methods differ in their way to obtain the maximum number of true hypotheses at each level. The Holm's procedure simply assigns the maximum number of true hypothesis at the stage $j$ to the number of remaining hypothesesat the $j^{th}$ stage, i.e. $t_j = k-j+1$.\\
In the Shaffer's method \cite{shaffer}, the possible numbers for true hypothesis and consequently $t_j$ is obtained by the following recursive formula
\begin{align*}
S(k) = \bigcup_{j=1}^k \{ \begin{pmatrix}2\\ j\\\end{pmatrix} + x: x \in S(k-j)  \}
\end{align*}
where $S(k)$ is the set of all possible numbers of true hypotheses when there are $k$ alignments for comparison and $S(0) = S(1) = {0}$. $t_j$ is simply computed based on the set $S(k)$.

Similar to the Shaffer's method, the Bergmann's method \cite{bergmann} use the logical interrelations between the hypotheses but dynamically estimates the maximum number of true hypotheses at the stage $j$, given that $j-1$ hypotheses are rejected.

To do so, they defined the exhaustive which is an index set of hypotheses $I \subseteq \{1,...,m \}$ where exactly all the hypotheses $H_j, j\in I$ can be true. For instance, let $A_1$, $A_2$, and $A_3$ be three alignments under study. If the null hypothesis between $A_1$ and $A_2$ is rejected, e.g., $A_1 \neq A_2$, then it is not possible that both hypothesis $A_1 = A_3$ and $A_2 = A_3$ be correct because the performance of $A_3$ cannot be the same as $A_1$ and $A_2$ while $A_1$ and $A_2$ have been already declared significantly different.

Having calculated the exhaustive set, any hypothesis $H_j$ is rejected if $j\notin A$ where $A$ is the acceptance set which is retained and defined as 
\begin{align}
A = \bigcup\{\text{I: I exhaustive}, \min\{P_i: i\in I\} > \alpha /|I| \}
\end{align}

The Bergmann's method is one of the most powerful procedures when $N\times N$ comparison is demanded since it dynamically takes into account the logical relations of hypothesis. However, building the exhaustive set is time-consuming especially if more than nine systems are available for comparison. 

\section{Results}
In this section, the recommended statistical procedures are applied to the OAEI 2016 \textit{anatomy} track, and the corresponding results are reported. Further, different string similarity metrics are compared and ranked according to the number of correct discoveries.

We have two ways of obtaining the contingency table, four McNemar's statistics and four ways to prevent FWER. Therefore, there are totally 32 states for comparison. On account of simplicity (and probably for the exclusion of duplication), we only consider four states: the two ways of building the contingency table compared with the McNemar's mid-p test and controlling FWER by the Nemenyi's and Bergmann's correction techniques, the most conservative and the most robust methods. The underlying reason behind the mid-p test selection is that it is not as conservative as the exact test and it is less likely to violate the nominal significant level $\alpha$ rather than the asymptotic test.

The anatomy track has been a part of OAEI since 2011 and its aim is to find the alignment between the Adult Mouse Anatomy and a part of the NCI Thesaurus related to the human anatomy. We select 10 systems participated in the OAEI 2016 for conducting the comparison: Alin \cite{alin}, AML \cite{aml}, CroMatcher \cite{cromatcher}, DKP-AOM \cite{dkpaom}, FCA-Map \cite{fcamap}, Lily \cite{lily}, LogMapLite \cite{logmap}, LPHOM \cite{lphom}, LYAM \cite{cromatcher}, and XMap \cite{xmap}. 

\begin{table}
\setlength\tabcolsep{6pt}
\caption{The $n_{01}$ and $n_{10}$ for constructing the contingency table from the first point of view which does not consider the false positives(see Eq. \eqref{ct_nfp}). For comparing the $i^{th}$ and $j^{th}$ systems, $n_{01} = (i,j)$ and $n_{10} = (j,i)$ where $(i,j)$ is the element at the $i^{th}$ row and the $j^{th}$ column in the table.} 
\label{tab:ct_nfp}
\begin{tabular}{r|ccccccccccc}
 & \rotatebox{90}{Alin} & \rotatebox{90}{AML} & \rotatebox{90}{CroMatcher} & \rotatebox{90}{DKP-AOM} & \rotatebox{90}{FCA\_Map} & \rotatebox{90}{Lily} & \rotatebox{90}{LogMapLite} & \rotatebox{90}{LPHOM} & \rotatebox{90}{LYAM} & \rotatebox{90}{XMap} \\ \hline
Alin & 0 & 0 & 13 & 405 & 2 & 18 & 2 & 52 & 3 & 0\\
AML & 911 & 0 & 62 & 1214 & 184 & 237 & 328 & 339 & 118 & 134\\
CroMatcher & 873 & 11 & 0 & 1170 & 176 & 216 & 311 & 314 & 108 & 124\\
DKP-AOM & 102 & 0 & 7 & 0 & 0 & 13 & 0 & 49 & 1 & 0\\
FCA\_Map & 763 & 34 & 77 & 1064 & 0 & 161 & 167 & 253 & 51 & 58\\
Lily & 713 & 21 & 51 & 1011& 95 & 0 & 176 & 210 & 45 & 60\\
LogMapLite & 597 & 12 & 46 & 898 & 1 & 76 & 0 & 203 & 5 & 19\\
LPHOM & 646 & 22 & 48 &  946 & 86 & 109 & 202 & 0 & 43 & 39\\
LYAM & 823 & 27 & 68 & 1124 & 110 & 170 & 230 & 269 & 0 & 74\\
XMap & 804 & 27 & 68 & 1107 & 101 & 169 & 228 & 249 & 58 & 0\\
\end{tabular}
\end{table}

\begin{table}
\setlength\tabcolsep{6pt}
\caption{The $n_{01}$ and $n_{10}$ for constructing the contingency table from the second point of view which takes into account the false positives (see Eq. \eqref{ct_wfp}). For comparing the $i^{th}$ and $j^{th}$ systems, $n_{01} = (i,j)$ and $n_{10} = (j,i)$ where $(i,j)$ is the element at the $i^{th}$ row and $j^{th}$ column in the table.}
\label{tab:ct_wfp}
\begin{tabular}{r|ccccccccccc}
 & \rotatebox{90}{Alin} & \rotatebox{90}{AML} & \rotatebox{90}{CroMatcher} & \rotatebox{90}{DKP-AOM} & \rotatebox{90}{FCA\_Map} & \rotatebox{90}{Lily} & \rotatebox{90}{LogMapLite} & \rotatebox{90}{LPHOM} & \rotatebox{90}{LYAM} & \rotatebox{90}{XMap} \\ \hline
Alin & 0 & 72 & 86 & 405 & 92 & 195 & 46 & 506 & 212 & 100\\
AML & 917 & 0 & 94 & 1214 & 252 & 396 & 368 & 777 & 298 & 203\\
CroMatcher & 879 & 42 & 0 & 1170 & 249 & 375 & 351 & 749 & 298 & 204\\
DKP-AOM & 108 & 72 & 80 & 0 & 90 & 190 & 50 & 509 & 210 & 100\\
FCA\_Map & 769 & 84 & 133 & 1064 & 0 & 323 & 181 & 691 & 220 & 135\\
Lily & 719 & 75 & 106 & 1011 & 170 & 0 & 219 & 617 & 234 & 138\\
LogMapLite & 597 & 74 & 109 & 898 & 55 & 246 & 0 & 648 & 186 & 107\\
LPHOM & 647 & 73 & 97 & 947 & 155 & 234 & 238 & 0 & 214 & 105\\
LYAM & 829 & 70 & 122 & 1124 & 160 & 327 & 252 & 690 & 0 & 142\\
XMap & 810 & 68 & 121 & 1107 & 168 & 324 & 266 & 674 & 235 & 0\\
\end{tabular}
\end{table}

The contingency table is built by two foregoing methodologies. The values of $n_{01}$ and $n_{10}$ for the first and second way of table construction are arranged in Tables \ref{tab:ct_nfp} and \ref{tab:ct_wfp}, respectively. For the interest of simplicity, $n_{01}$ and $n_{10}$ are tabulated in one single table for each perspective (below and upper diagonal). To compare the $i^{th}$ and $j^{th}$ systems in each approach, $(i,j)$ and $(j,i)$ elements of this table are taken as $n_{01}$ and $n_{10}$, where $(i,j)$ is the element at the $i^{th}$ row and $j^{th}$ column. For instance, let's compare $Alin$ and $AML$ systems. In the first perspective, $n_{01} = 911$ which means that there are $911$ correspondences discovered by AML but not by Alin. And, $n_{10} = 0$ indicates that there are no correspondences identified by Alin but not by AML. In the second perspective, on the other hands, $n_{01} = 917$ and $n_{10} = 72$. Comparing with the previous view, $n_{10}$ changes from 0 to 72 which means that AML has discovered 72 wrong correspondences while Alin has not. The little increase in $n_{01}$ is due to the false discovery rate of Alin (6 correspondences) in comparison to AML. As a result, it is grasped that the false discovery rate of Alin is less than AML while the true discovery rate of AML is way higher than Alin. If the McNemar's test rejects the null hypothesis, AML is thus concluded to have a better performance than Alin due to its higher true discovery rate. The comparison of other systems can be conducted likewise which clarifies the difference between two perspectives.

\begin{figure}[h!]
  \caption{Comparison of alignment systems by the McNemar's mid-p test with the Nemenyi's correction while the false positive is ignored. The edge $A \rightarrow B$ indicates that A outperforms B. }
  \includegraphics[scale=.49]{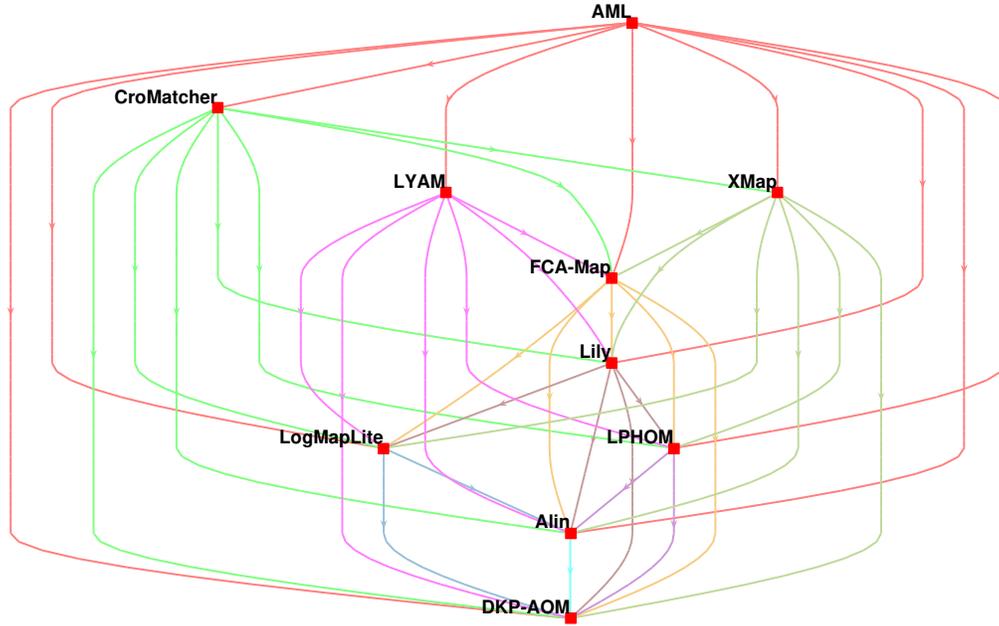}
  \label{fig:nfp_nemenyi}
\end{figure}

\begin{figure}[h!]
  \caption{Comparison of alignment systems by the McNemar's mid-p test with the Bergmann's correction while the false positive is ignored. The edge $A \rightarrow B$ indicates that A outperforms B.}
  \includegraphics[scale=.48]{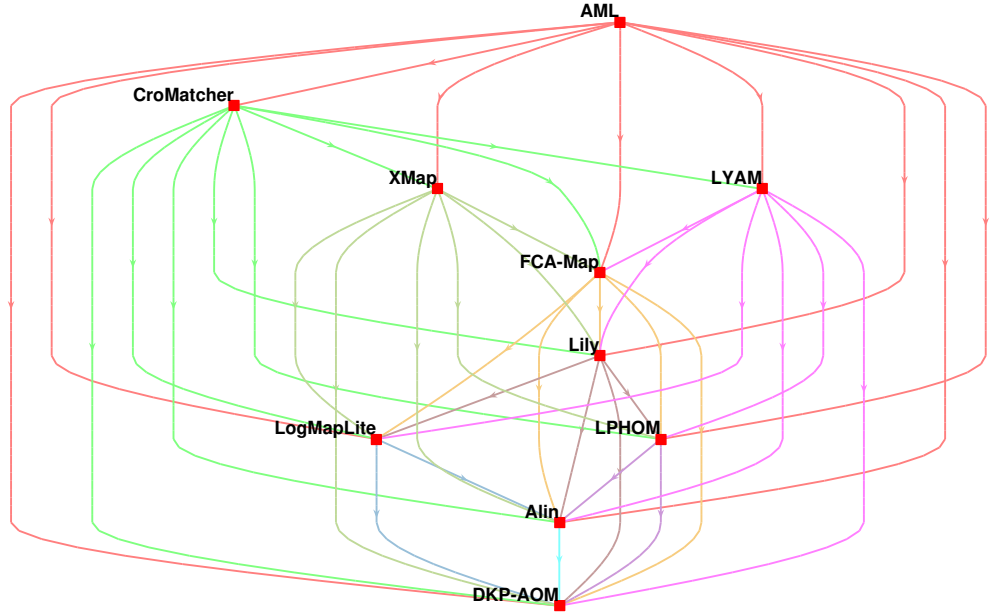}
  \label{fig:nfp_bergmann}
\end{figure}

\begin{figure}[h!]
  \caption{Comparison of alignment systems by the McNemar's mid-p test with the Nemenyi's correction while the false positive is considered. The edge $A \rightarrow B$ indicates that A outperforms B.}
  \includegraphics[scale=.49]{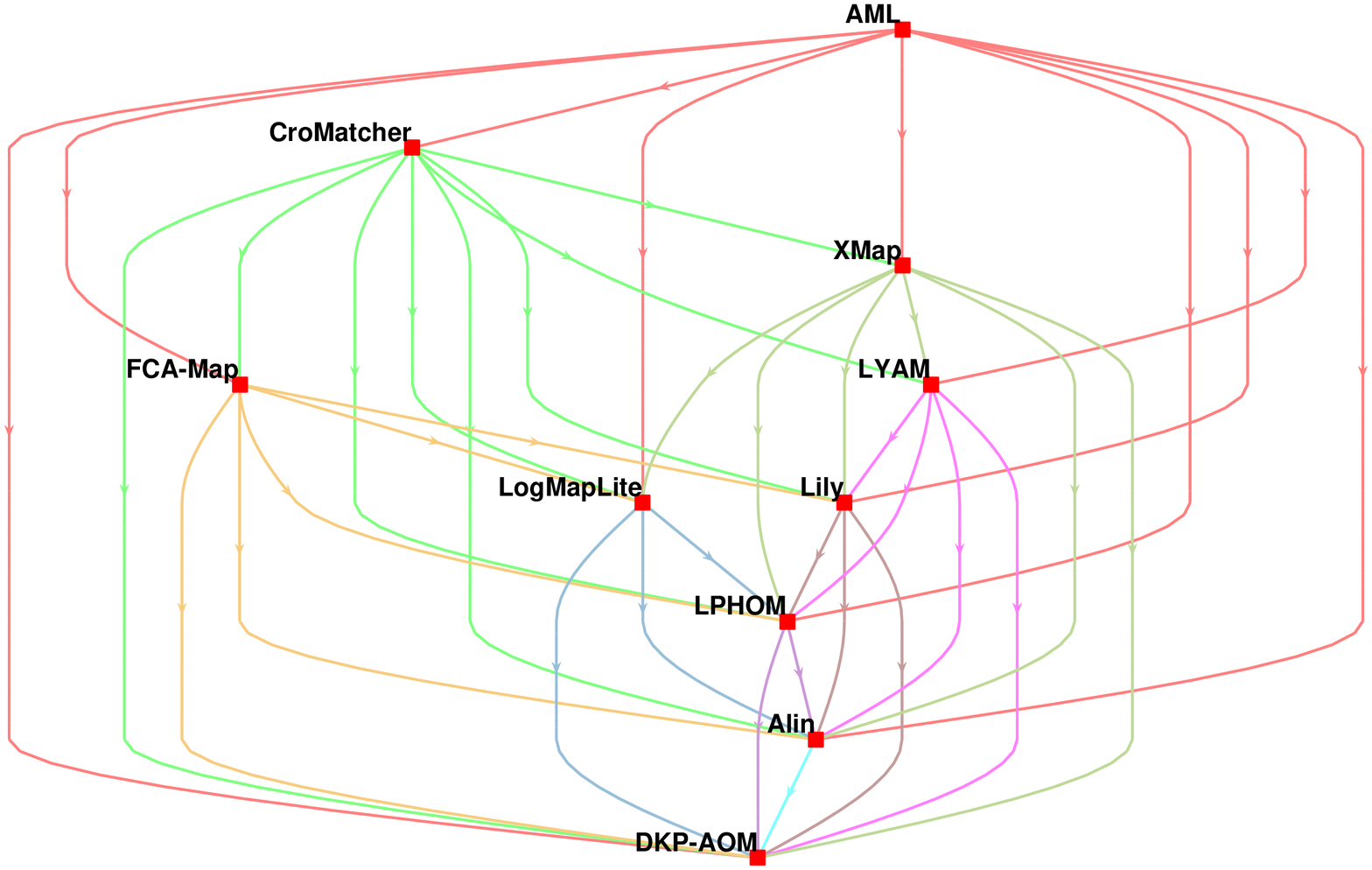}
  \label{fig:wfp_nemenyi}
\end{figure}

\begin{figure}[h!]
  \caption{Comparison of alignment systems by the McNemar's mid-p test with the Bergmann's correction while the false positive is considered. The edge $A \rightarrow B$ indicates that A outperforms B.}
	\includegraphics[scale=.49]{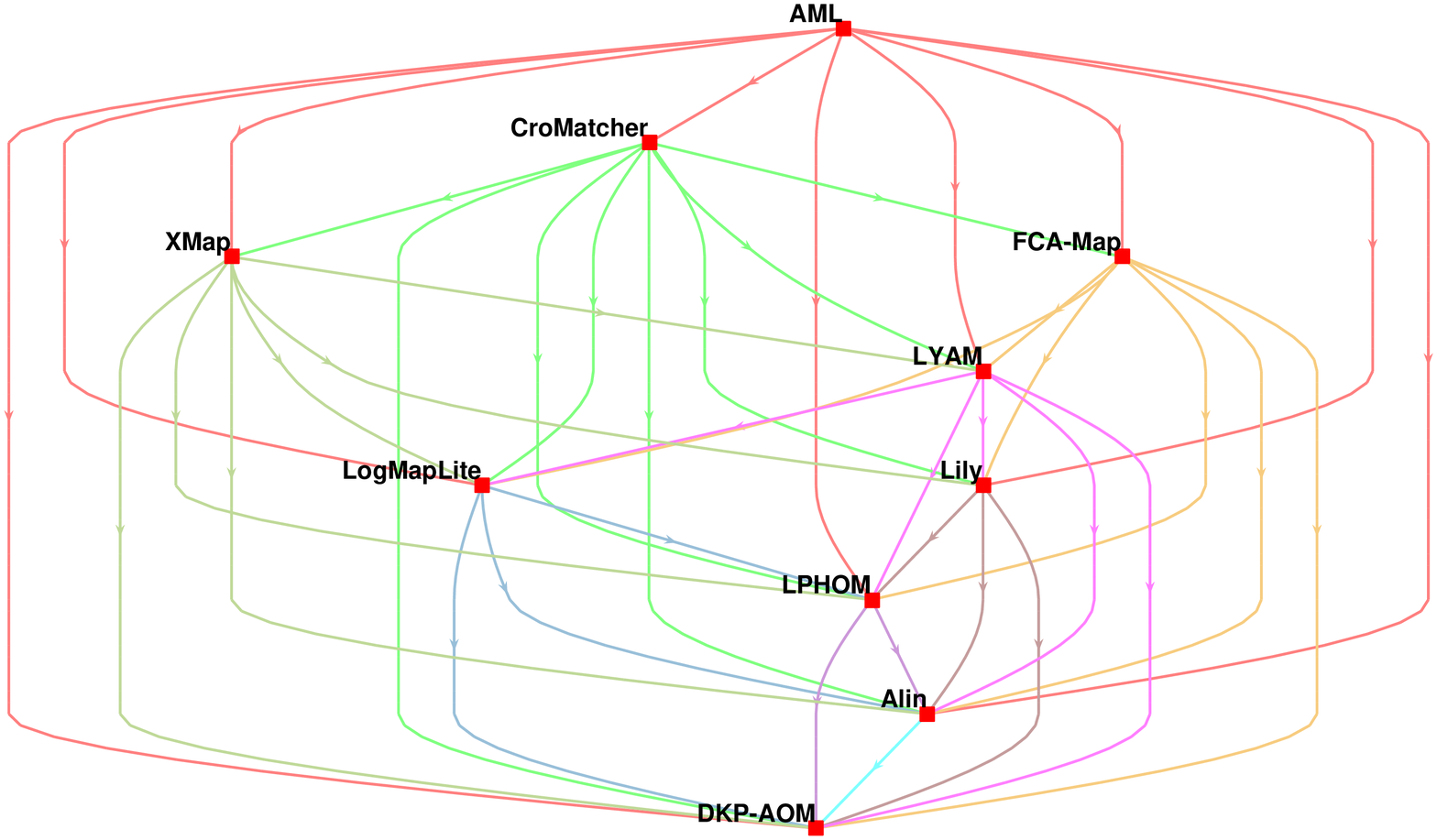}
  \label{fig:wfp_bergmann}
\end{figure}

We conduct all the pairwise comparisons and we take advantage of the Nemenyi's correction and the Bergman's correction, the most conservative and most powerful ones, to control the family-wise error rate. A directed graph visualizes the outcome of the pairwise comparison. Four different directed graphs correspond to each perspective and each correction method are displayed in Figures (\ref{fig:nfp_nemenyi} - \ref{fig:wfp_bergmann}). The nodes in these graphs are the systems under study and any directed edge $A \rightarrow B$ means that A is significantly better than B. If there is no such an edge, however, there is no significant difference between the corresponding systems.

First, we compare the results obtained from the Nemenyi's and Bergman's correction techniques from each perspective of the contingency table construction. Figures \ref{fig:nfp_nemenyi} and \ref{fig:nfp_bergmann} are the directed graphs corresponds to the pairwise comparisons of alignments obtained by applying respectively the Nemenyi's and Bergmann's correction under the first perspective of contingency table construction. The results of these two correction methods are varied only in one comparison: the Bergmann's correction indicates the significant difference between CroMatcher and LYAM while the Nemenyi's correction cannot detect it. Thus, the Bergmann's correction is more powerful than the Nemenyi's method as the theory suggests.

In the second approach which considers the false positive, the Bergmann's correction indicates its power in comparison with the Nemenyi's correction. It declares the difference between FCA-Map and LYAM, and between LYAM and LogMapLite significant while the Nemenyi's correction cannot find such differences as significant.

Now, we compare the two perspectives on the contingency table construction. To do so, the Bergmann's correction method is considered due to its ability to detect more differences. Considering Fig. \ref{fig:nfp_bergmann}, it is readily seen that the LYAM and XMAP methods are not declared significant, but both of them are declared significant in comparison to FCA-MAP. If the false positive rate is taken into account, as in Fig. \ref{fig:wfp_bergmann}, FCA-MAP is replaced LYAM. To investigate such a replacement, Tables \ref{tab:ct_nfp} and \ref{tab:ct_wfp} must be considered. While the false positive rate is not considered, FCA-Map has 51 correct correspondences which are not in LYAM, and LYAM has 110 true correspondences which do not exist in FCA-MAP. However, when the false positive is considered, the number of truly discovered correspondences by FCA-MAP which are not in the LYAM alignment increases to 220 while the number of truly discovered correspondences by LYAM which are not in FCA-MAP is 160. As a result, the LYAM ontology mapping is better than FCA-MAP from the first point of view, but FCA-MAP outperforms LYAM in the second approach because it has a lower false discovery rate in comparison with LYAM. The same argument is also valid for the comparison of FCA-MAP and XAMP: if the falsely discovered correspondences are not taken into account, XAMP outperforms FCA-MAP while they are declared insignificant when the false discovery error is considered as well. 

Another difference between two perspectives on the contingency table construction is about the LogMapLite system. When the false discovery rate does not matter, Lily outperforms LogMapLite, which is further declared insignificant compared with LPHOM. If the false positive error is heeded, however, LogMapLite outperforms LPHOM and it is declared insignificant with Lily. This indicates that LogMapLite has a lower false discovery rate than Lily and LPHOM.

We rank the systems participated in the OAEI 2016 anatomy track in Table \ref{tab:rank} based on the Bergmann's correction. The columns with labels IFP and CFP correspond to the contingency table construction with ignoring the false discovery (IFP) and considering (CFP) it. In this table, the systems in higher rows are ones which are significantly better than the ones in the lower rows. If two systems are not significantly different, they are placed in the same cell. It can be readily seen that AML and DKP-AOM are the best and the worst systems from two perspectives, respectively.

The results of statistical procedures are eventually compared with those of recall and F-measure. As a matter of fact, such a comparison would be of no meaning unless some circumstances would be considered. We say that two systems are not significantly different provided that their recall (or F-measure for another case) will be the same. Nonetheless, it must be mentioned that the comparison based on the McNemar's test is distinct from that of different performance measures. First and foremost, it does not produce any score. Second, the result of comparison might indicate that two systems are similar, the case which is not accommodated in comparison of two scores unless they are exactly the same.

First, the outcomes of our analysis from the first perspective with the Bergmann's correction (see Figure \ref{fig:nfp_bergmann}) is compared with the recall metric. In the OAEI 2016 anatomy track, AML and CroMatcher have the highest recall among others. At the other extreme, DKP-AOM and Alin are the systems with the least discovery. By the same token, they are the top two and bottom two systems in our analysis. One salient characteristic of the statistical analysis is the equivalence of LPHOM and LogMapLite. The recall of LogMapLite and LPHOM are 0.728 and 0.727, respectively. If the higher recall would be an indicator for superiority, then LogMapLite is declared better. However, the difference between these systems is a trifle. This triviality is reflected in the statistical analysis as they are not declared significant (there is no edge between LogMapLite and LPHOM in Figure \ref{fig:nfp_bergmann}). There is the same cogent argument for the comparison of XMap and LYAM.

The comparison of the second perspective is analogous to that of the F-measure. Similar to our analysis, the F-measures of AML and CroMatcher are the top systems, and those of DKP-AOM and Alin are the bottom two ones (see Figure \ref{fig:wfp_bergmann}).

\begin{table}
\caption{Ranking of methods participated in the anatomy track, OAEI 2016 from two different perspectives. The first perspective is to ignore the false positive (IFP) and the second is to consider it (CFP). The position of upper rows in this table indicates that it is significantly better than the methods coming in the lower rows. Cells with two methods indicate that the methods are not declared significantly different.}
\label{tab:rank}
\begin{tabular}{c | c | c}
&IFP 		& 		CFP \\ 
\hline
1 &AML    & AML \\
2 &CroMatcher  & CroMatcher \\
3 &LYAM $\&$ XAMP &FCA-MAP $\&$ XMAP \\
4 &FCA-MAP & LYAM \\
5 &Lily & LogMapLite $\&$ Lily \\
6 &LogMapLite $\&$ LPHOM & LPHOM \\
7 &Alin & Alin \\
8 &DKP-AOM & DKP-AOM
\end{tabular}
\end{table}

For the final experiment, the string-based similarity measures are compared over the anatomy track. These metrics are of utmost importance, by which most of the correspondences of two given ontologies, including the ontologies of the anatomy track, could be discovered \cite{michelle}. To compare such metrics over the anatomy track, we take advantage of the Shiva framework \cite{shiva} which converts the ontology mapping into an assignment problem. In this framework, the similarity between each concept from the source ontology is gauged with all the concepts of the target ontology. The similarity score between the concepts of two ontologies constructs a matrix, which can be given to the Hungarian algorithm \cite{hungarian} to find the best match for each entity. We use nine string-based similarity measures to construct the matrix: Levenstein \cite{levenstein}, N-gram \cite{ngram}, Hamming \cite{om_book}, Jaro \cite{jaro}, JaroWinkler \cite{jarowinkler}, SMOA \cite{smoa}, NeedlemanWunsch2 \cite{nw2}, Substring distance \cite{om_book}, and equivalence measure. The Hungarian method applies to the resultant matrix to find the best match for each concept.
\begin{table}
\caption{The $n_{01}$ and $n_{10}$ for constructing the contingency table from the first point of view (ignoring the false positive) across the various string-based similarity measures. For the comparison of the $i^{th}$ and $j^{th}$ metrics, $n_{01} = (i,j)$ and $n_{10} = (j,i)$ where $(i,j)$ is the element at the $i^{th}$ row and the $j^{th}$ column in the table.}
\begin{tabular}{r|ccccccccc}
 & \rotatebox{90}{Equal} & \rotatebox{90}{Hamming} & \rotatebox{90}{Jaro} & \rotatebox{90}{JaroWinkler} & \rotatebox{90}{Levenshtein} & \rotatebox{90}{N-gram} & \rotatebox{90}{Needleman.} & \rotatebox{90}{SMOA} & \rotatebox{90}{SubString} \\ \hline
Equal & 0 & 0 & 2 & 2 & 0 & 0 & 0 & 71 & 0\\
Hamming & 842 & 0 & 51 & 51 & 32 & 54 & 48 & 258 & 494\\
Jaro & 888 & 95 & 0 & 0 & 42 & 59 & 60 & 252 & 532\\
JaroWinkler & 888 & 95 & 0 & 0 & 42 & 59 & 60 & 252 & 532\\
Levenshtein & 966 & 156 & 122 & 122 & 0 & 64 & 50 & 277 & 593\\
N-gram & 1041 & 253 & 214 & 214 & 139 & 0 & 174 & 290 & 636\\
Needleman. & 932 & 138 & 106 & 106 & 16 & 65 & 0 & 276 & 573\\
SMOA & 880 & 225 & 175 & 175 & 120 & 58 & 153 & 0 & 552\\
SubString & 422 & 74 & 68 & 68 & 49 & 17 & 63 & 165 & 0\\
\end{tabular}
\label{tab:anatomy_str}
\end{table}

\begin{figure}[t]
\centering
\caption{comparison of string-based similarity measures for the anatomy track. The arrow $A \rightarrow B$ indicates that A outperforms B.}
\includegraphics[scale=.7]{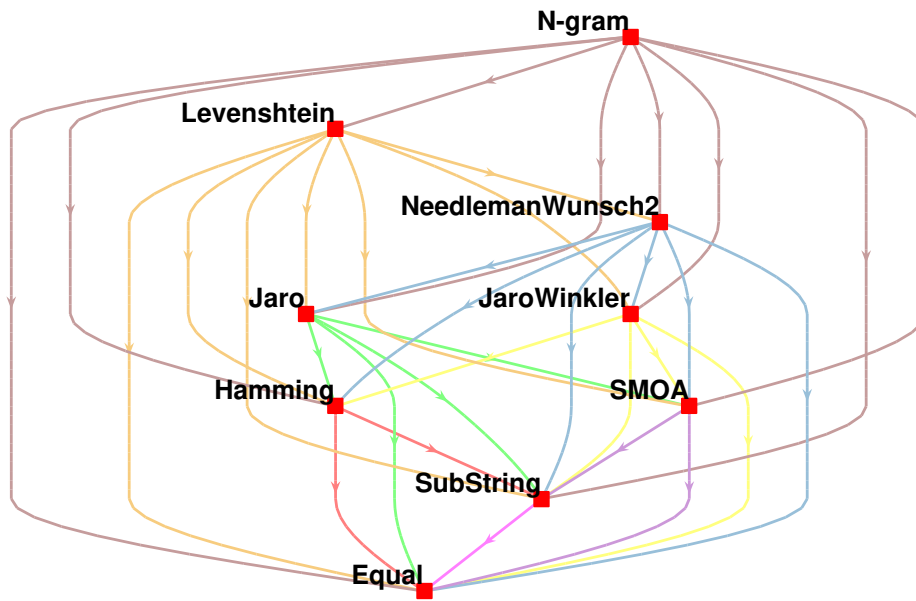}
\label{fig:anatomy_str}
\end{figure}
We consider the case when the false positive is not taken into account. The primary reason is that the selection of the appropriate string similarity measure can enable us to discover most of the potential correspondences \cite{michelle}. If the right similarity metric is chosen, then the unreliable correspondences could be omitted by applying more strict thresholds.
\\

Similar to the previous ones, Table \ref{tab:anatomy_str} tabulates $n_{01}$ and $n_{10}$ corresponding to different string-based similarity measures while the false positive is ignored. The results are visualized by a directed graph shown in Fig. \ref{fig:anatomy_str}. From this figure, N-gram has shown the best performances and is followed by Levenstein. Further, SMOA and Hamming distances are the ones with the least retrieved correspondences but they are better than Substring and Equivalence measures as expected.

\section{Conclusion}
This paper proposed the utilization of the McNemar's test to compare various ontology alignment systems over one single task. The current approach for the alignment comparison is to first select a performance score and then compare two systems by obtaining their performance scores on a task with a reference alignment. In this article, the alignment produced by two systems as well as the reference alignment are given, and the outcome is if two systems are significantly different. Thus, the output is not a score, but to / not to declare the significance between two ontology matching technique. Further, the ways of preventing family-wise error rate, which is likely to happen in the comparison of multiple ($>2$) alignment systems, are explored in minute detail. The proposed methodologies are applied to the anatomy track of ontology alignment initiative evaluation (OAEI) 2016. It is indicated that the AML and CroMatcher are the top two algorithms, and Alin and DKP-AOM are the worst alignments. For string-based similarity measures, N-gram and Levenstein outperform other methods while SMOA and Hamming distance have shown poor performances. 

\bibliographystyle{ACM-Reference-Format} 
\bibliography{ref}

\end{document}